 \documentclass[pmlr,twocolumn]{jmlr} 

\usepackage{booktabs}

\usepackage[load-configurations=version-1]{siunitx} 
\usepackage{caption}

 \usepackage{siunitx}
\usepackage{mathptmx}
 \usepackage{mathtools}
\usepackage{booktabs, multirow}

\usepackage{graphicx}
\theorembodyfont{\upshape}
\theoremheaderfont{\scshape}
\theorempostheader{:}
\theoremsep{\newline}

\jmlryear{2020}
\jmlrworkshop{Preprint-version} 

\title[Pay Attention to the Cough]{Pay Attention to the cough: \\ Early Diagnosis of COVID-19 using Interpretable Symptoms Embeddings with Cough Sound Signal Processing}

\author{%
\Name{Ankit Pal} \Email{ankit.pal@saama.com}\\
\addr Saama AI Research, Chennai, India
\AND
\Name{Malaikannan Sankarasubbu} \Email{malaikannan.sankarasubbu@saama.com}\\
\addr Saama AI Research, Chennai, India
}

\begin{document}
\maketitle
\begin{abstract}
COVID-19 (coronavirus disease 2019) pandemic caused by SARS-CoV-2 has led to a treacherous and devastating catastrophe for humanity. At the time of writing, no specific antivirus drugs or vaccines are recommended to control infection transmission and spread. The current diagnosis of COVID-19 is done by Reverse-Transcription Polymer Chain Reaction (RT-PCR) testing. However, this method is expensive, time-consuming, and not easily available in straitened regions.
An interpretable and COVID-19 diagnosis AI framework is devised and developed based on the cough sounds features and symptoms metadata to overcome these limitations. The proposed framework's performance was evaluated using a medical dataset containing Symptoms and Demographic data of 30000 audio segments, 328 cough sounds from 150 patients with four cough classes ( COVID-19, Asthma, Bronchitis, and Healthy). Experiments' results show that the model captures the better and robust feature embedding to distinguish between COVID-19 patient coughs and several types of non-COVID-19 coughs with higher specificity and accuracy of 95.04 $\pm$ 0.18 \% and 96.83$ \pm$ 0.18 \% respectively, all the while maintaining interpretability.
\end{abstract}
\begin{keywords}
COVID-19, Audio Analysis, Deep Learning, Medical Data, Machine Learning
\end{keywords}

\section{Introduction} \label{sec:intro}
The novel coronavirus (COVID-19) disease has affected over 31.2 million lives, claiming more than 1.02 million fatalities globally, representing an epoch-making global crisis in health care. At the time of writing, no specific antivirus drugs or vaccines are recommended to control transmission and spread infection. The current diagnosis of COVID-19 is made by Reverse-Transcription Polymer Chain Reaction (RT-PCR) testing, which utilizes several primer-probe sets depending on the assay utilized \citep{Emery2004}. However, this method is time-consuming, expensive, and not easily available in straitened regions due to lack of adequate supplies, healthcare facilities, and medical professionals. A low-cost, rapid, and an easily accessible testing solution is needed to increase the diagnostic capability and devise a treatment plan.
Computed Tomography(CT) helps clinicians perform complete patient assessments and describe the specific characteristic manifestations in the lungs associated with COVID-19 \citep{Li2020}. Hence, serving as an efficient tool for early screening and diagnosis of COVID-19. 
In analyzing medical images, AI-based methods have shown great success \citep{Du2018, Heidari2018, Heidari2020}. These methods are scalable, automatable, and easy to apply in clinical environments \citep{Ahmed2020, Shah2019}. Significant attempts have been made to use x-ray images for automatic diagnosis of COVID-19 \citep{Pereira2020, Narin2020, Zhang2020, Apostolopoulos2020}. Studies dealing with the classification of COVID-19 show promising results in this task.

However, in \citep{Cohen2020} work, the classification limitations of x-ray images are examined since the network may learn more unique features to the dataset than those unique to the disease. Despite its success, the CT scan displays similar imaging characteristics, making it difficult to distinguish between COVID-19 and other pneumonia types. Moreover, CT-based methods can be integrated only with the Healthcare system to help clinical doctors, radiologists, and specialists detect COVID-19 patients using chest CT images. Unfortunately, an individual cannot utilize this method at home. To obtain the CT scan image and report, one must visit a well-equipped clinical facility or diagnostic center, which may increase the risk of exposure to the virus.
According to the WHO and CDC official report, the four primary symptoms of the COVID are dry cough, fever, tiredness, and difficulty in breathing. \citep{CDC2019}. However, cough is more common as it is one of the early symptoms of respiratory tract infections. Studies show that it occurs in 68\% to 83\% of the people showing up for the medical examination. Cough classification is usually carried out manually during a physical examination, and the clinician may listen to several episodes of voluntary or natural coughs to classify them. This information is crucial in diagnosis and treatment.

In previous studies, several methods with speech features have been proposed to automate different cough types classification. In the study published by \citep{Knocikova2008}, the sound of voluntary cough in patients with respiratory diseases was investigated. Later, in 2015, \citep{Guclu2015} published the study on the analysis of asthmatic breathing sounds. These studies utilized the wavelet transformation, which is a type of signal processing technique, generally used on non-stationary signals. In a study by \citep{Swarnkar2012}, a Logistic Regression model was utilized to classify the dry and wet cough from pediatric patients with different respiratory illnesses. For pertussis cough classification, three separate classifiers' performance was analyzed in \citep{Parker2013} research.
Several AI-based approaches, motivated by prior work, have been presented to detect patients with COVID-19 using cough sound analysis. \citep{Deshpande2020} gives an overview of Audio, Signal, Speech, NLP for COVID-19, \citep{Orlandic2020, Brown2020, Sharma2020} have collected a crowdsourced dataset of respiratory sounds and shared the findings over a subset of the dataset. \cite{Imran2020, Furman2020} performed similar analyses on cough data and achieved good accuracy. Most studies use short-term magnitude spectrograms transformed from cough sound data to the convolutional neural network (CNN).
However, these methods have the following limitations :
\begin{itemize}
\item \textbf{Ignoring domain-specific sound information} Cough is a non-stationary acoustic event. CNN is based only on a spectrogram input; some domain-specific important characteristics (besides spectrogram) of cough sounds might be overlooked in the feature space.
\item \textbf{Using cough features only} These methods exploit the cough features only, ignoring patient characteristics, medical conditions, and symptoms data. Both cough features and other symptoms accompanied by demographic data are responsible for COVID-19 infection. Wherein the prior carries vital information about the respiratory system and the pathologies involved, the latter encodes patient characteristics, signs, and health conditions (fever, chest pain, dyspnea). However, their existence alone is not a precise enough marker of the disease. Therefore, determining the symptoms (besides cough) presented by suspected cases, as best predictors of a positive diagnosis would be useful to make rapid decisions on treatment and isolation needs.
\item \textbf{Lack of interpretability} In AI research, the model is not limited to accuracy and sensitivity reports; instead, it is expected to describe the predictions' underlying reasons and enhance medical understanding and knowledge. Clinical selection of an algorithm depends on two main factors, its clinical usefulness, and trustworthiness. When the prediction does not directly explain a particular clinical question, its use is limited.
\end{itemize}
To overcome the limitation of the existing methods, A novel interpretable COVID-19 diagnosis AI framework is proposed in this study, which uses symptoms and cough features to classify the COVID-19 cases from non-COVID-19 cases accurately. A three-layer Deep Neural Network model is used to generate cough embeddings from the handcrafted signal processing features and symptoms embeddings are generated by a transformer-based self-attention network called TabNet. \cite{arik2020tabnet} Finally, the prediction score is obtained by concatenating the Symptoms Embeddings with Cough Embeddings, followed by a Fully Connected layer.
In a sensitive discipline such as healthcare, where any decision comes with an extended and long term responsibility, making wrong predictions can lead to critical judgments in life and death situations.

In this study, it is illustrated that this framework is not limited to accurate predictions or projections. Instead, it explains the underlying reasons for the same and answers the question as to why the model predicts it. The contributions of the paper can be summarized as follows:
\begin{itemize}
\item A novel explainable \& interpretable COVID-19 diagnosis framework based on deep learning (AI) uses the information from symptoms and cough signal processing features.
The proposed solution is a low-cost, rapid, and easily accessible testing solution to increase the diagnostic capability and devise a treatment plan in areas where adequate supplies, healthcare facilities, and medical professionals are not available.

\item In this study, an interpretable diagnosis solution is presented, capable of explaining and establishing a dialogue with its end-users about the underlying process. Hence, resulting in transparent human interpretable outputs. 
\item Three binary and one multi-class classification tasks are developed in this study; Task 1 uses only cough features to classify between COVID-19 positive and COVID-19 negative. In Task 2, only demographic and symptoms data is used, and in Task 3, both types of information are used, which helps the model learn deeper relationships between temporal acoustic characteristics of cough sounds and Symptoms' features and hence perform better. In Task 4, multi-class classification is performed to explain the proposed model's effectiveness in classifying between four cough types, including Bronchitis, Asthma, COVID-19 Positive, and COVID-19 Negative.
\item An in-depth analysis is performed for different cough sounds. The observations and findings are presented, distinguishing COVID-19 cough from other types of cough. 
\item A python module was developed to extract better and re-boost cough features from raw cough sounds. This module is open-sourced to help users, developers, and researchers. Those are not necessarily experts in domain-specific cough feature extraction, contributing to real-time cough based research application, and provide better mobile health solutions.
\item This study hence provides a medically-vetted approach.
\end{itemize}
\section{Model Architecture}
The model architecture consists of two subnetworks components, including the Symptoms Embedddings and Cough Embeddings , that process the data from different modalities.
\subsection{Symptoms Embeddings}
Symptoms Embeddings capture the hidden features of patient characteristics, diagnosis, symptoms.
\begin{figure*}[!h]
  \includegraphics[width=15cm]{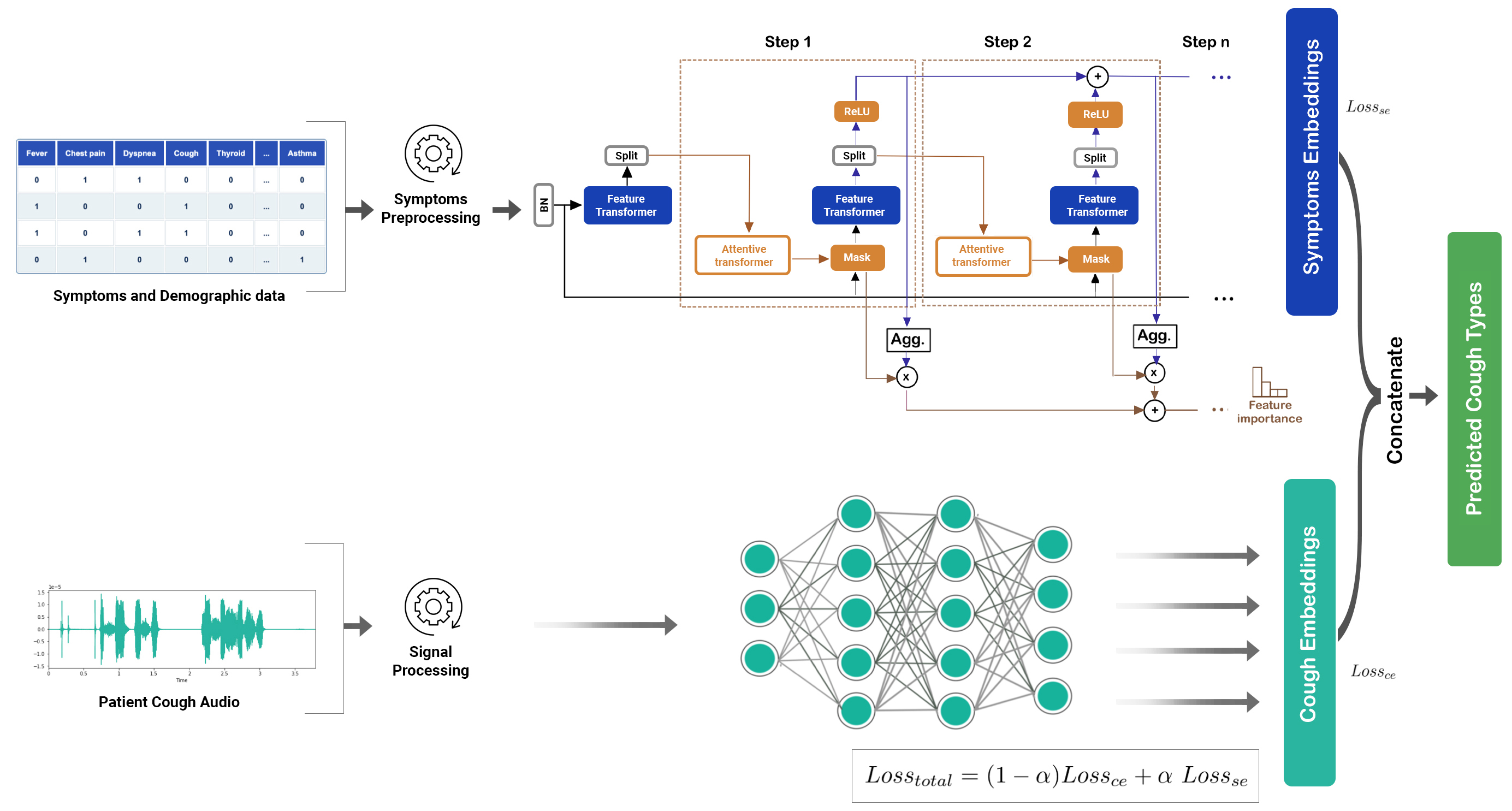}
  \caption{ \footnotesize Illustration of the overall structure of Proposed Model, The model consists of two subnetworks that process the data from different modalities. The TabNet network at the top, process the features that include symptoms \& Demographic data, while the network in the bottom processes the audio signal from cough. Attention Masks from each decision step are aggregated to provide model interpretability results}
  \label{fig:maindiagram}
\end{figure*}
A feature that has been masked a lot has low importance for the model and vice-versa. Averaged Attention masks are used to explain the overall importance of symptoms features.
\subsubsection{Decision Step (DS)}
TabNet stacks the subsequent DS one after the other. 
Decision steps are composed of a Feature Transformer(FT) \appendixref{apd:ft}, an Attentive Transformer(AT)\appendixref{apd:at} and feature masking. Symptoms features are mapped into a D-dimensional trainable embeddings $q \in \mathbb{R}^{B \times D}$, where $B$ is the batch size, and $D$ is the feature dimension. A batch normalization (BN) is performed across the whole batch. For the selection of specific soft features and explain the feature importance  TabNet uses a learnable mask
$M[i] \in \mathbb{R}^{B \times D}$.
Each decision step has a specific mask and selects its own features; steps are sequential, so the second step needs the first to be finished. 

We obtain the mask(M) output at each decision step  by multiplying the mask with Normalised symptoms features $q_{i}\bigl( M[i]\cdot q \bigr)$ 
Normalized domain features are passed to FT, and a split block divides the processed representation into two chunks for the next decision step.

\begin{equation}
\bigl[d[i], a[i]\bigr] = q_{i}(M[i]\cdot q)
\end{equation}

where $d[i] \in \mathbb{R}^{B \times n_{d}}$ and $a[i] \in \mathbb{R}^{B \times n_{a}}$.
and $n_{d}$, $n_{a}$ are size of the decision layer and size of the attention bottleneck respectively.
After $n^{th}$ step two outputs are produced.
\begin{itemize}
\item Mask outputs are aggregated from all the decision steps to provide model interpretability result. \figureref{fig:att0}  and \figureref{fig:att1} shows the interpretability result.
\item The final output is a linear combination of the all the summed decision steps, similar to the decision tree result.
\begin{equation}
\mathrm{d}_{\mathrm{o u t}}=\sum_{i=1}^{N_{s t e p s}} \operatorname{ReLU}(d[i]) \end{equation}
\end{itemize}
Pairwise dot product was computed between output $\mathrm{d}_{\mathrm{o u t}}$ and  FC layer to obtain the Symptoms Embeddings $\mathbf{S}_{e} \in \mathbb{R}^{B \times F}$ where $B$ is the batch size  and $F$ is the output dimension.
\subsubsection{ Symptoms Embedding Loss}
TabNet uses regularized sparse entropy loss to control the sparsity of attentive features. The regularization factor is a mathematical aggregation of the attention mask.

\begin{equation}
\mathrm{Loss}_{se}=
\smashoperator{\sum_{i=1}^{N_{\mathrm{steps}}}}
\sum_{b=1}^{B} \sum_{j=1}^{D} 
\frac{-M_{b,\mkern1.5mu j}[i]}{%
   N_{steps}\, B} 
\log (M_{b,\mkern1.5mu j}[i]+\epsilon)
\end{equation}

Where $\epsilon$ is a small positive value.

\subsection{Cough Embeddings}
Cough Embeddigs learn and capture deeper features in temporal acoustic characteristics of cough sounds.
\subsubsection{Signal Preprocessing}
\label{sec:sp}

Before extracting cough features and feeding it to Deep Neural Networks(DNN), some pre-processing of raw audio data is needed. Each cough recording was downsampled to 16 kHz; normalization was applied to the cough signal level with a target amplitude of -28.0 dBFS to keep the future features as close as possible to the same level.

Normalized features were split into cough segments based on the silence threshold. Let $s[t]$ be the discrete-time cough sound recording. 
The expression of signal $s[t]$ can be written as:
\begin{equation}
s[t] = y[t] + b[t]
\end{equation}

Where $y[t]$ denotes the cough signal $b[t]$ denotes the noise in the signal. 
To reduce the noise and get $y[t]$ a High Pass Filter(HPF) was applied on $s[t]$. Cough segments $y[t]$ were divided into sub segments of non-overlapping Hamming-windowed frames. Let  $y_{i}[t], i = 1,2,3,…n$ denotes the $i^{th}$ cough sub-segment from signal $y[t]$ with length N.

A total of 22 cough features were extracted from each cough recording segment $y[t]$, including 12 MFCC features, 4 Formant frequencies, Zcr, Kurtosis feature, Log energy, Skewness feature, Entropy, and Fundamental frequency(F0).
Please see \appendixref{apd:cfeatures} for cough features. Also see \appendixref{apd:dataacq} for detailed information about our data collection process.
The final feature matrix was grouped by chunks of $n$ Consecutive feature matrix, and A total of 44 cough features were extracted by taking the mean and standard deviation for all the cough features in each chunked matrix.

Later we feed the final feature matrix to 3 layers Deep Neural Network(DNN) with ReLu activation function to get the final Cough Embeddings  $\mathbf{C}_{e} \in \mathbb{R}^{B \times F}$ where $B$ is the batch size  and $F$ is the output dimension.

\begin{table*}
\centering
\begin{tabular}{c|ccccc}
    \toprule
    \midrule
        & \textbf{F1-score}  &  \textbf{Precision}  & \textbf{Sensitivity}  & \textbf{Specificity}  & \textbf{Accuracy}  \\
    \midrule
    
        Covid-19 Positive &\footnotesize 86.38 $\pm$ 0.03\%     & \footnotesize 81.88 $\pm$ 0.01\% & \footnotesize 91.39 $\pm$ 0.04\% & \footnotesize 97.49 $\pm$ 0.03\% & \footnotesize 96.81 $\pm$ 0.05\% \\
        Covid-19 Negative &\footnotesize 92.16 $\pm$ 0.01\%  &\footnotesize  95.09 $\pm$ 0.02\% &\footnotesize  89.41 $\pm$ 0.08\% & \footnotesize 98.64 $\pm$ 0.05\%& \footnotesize 96.55 $\pm$ 0.11\%             \\
        
        Bronchitis &\footnotesize 92.85 $\pm$ 0.04\% & \footnotesize 97.70 $\pm$ 0.05\% & \footnotesize 88.45 $\pm$ 0.05\% & \footnotesize 98.08 $\pm$ 0.12\% & \footnotesize 93.46 $\pm$ 0.02\%  \\
        
        Asthma &\footnotesize 83.88 $\pm$ 0.13\%  & \footnotesize 75.46 $\pm$ 0.03\% & \footnotesize 94.41 $\pm$ 0.04\% & \footnotesize 93.10 $\pm$ 0.01\% & \footnotesize 93.34 $\pm$ 0.05\% \\
        \midrule
        Overall & \footnotesize 90.09 $\pm$ 0.17\% & \footnotesize 90.92 $\pm$ 0.09\% & \footnotesize 90.41 $\pm$ 0.14\% & \footnotesize 96.83 $\pm$ 0.06\% & \footnotesize 95.04 $\pm$ 0.18\% 
        
        \\
 \bottomrule
\end{tabular}
\caption{Model performance metrics across four different diseases}
\label{tab:fourdiseases}
\end{table*}
\subsubsection{Cough Embeddings Loss}
In Multi-class classification setting, we use Categorical Crossentropy loss function to calculate the loss of Cough Embeddings
\begin{equation}
\mathrm{Loss}_{\mathrm{ce}} =-\sum_{i=1}^{\operatorname{N}} y_{i} \cdot \log \hat{y}_{i}
\end{equation}

where N is the number of classes in dataset, $\hat{y}_{i} $ denotes the $\textit{i}$-th predicted class in the model output and $y_{i}$ is the corresponding target value.
In a binary setting, we use the Binary Cross-Entropy loss function.

\begin{equation}
\mathrm{Loss}_{\mathrm{ce}} = -\frac{1}{N} \sum_{n=1}^{N}\left[y_{n} \log \hat{y}_{n}+\left(1-y_{n}\right) \log \left(1-\hat{y}_{n}\right)\right]
\end{equation}
\subsection{Classification layer}
We get the prediction score by concatenating the Symptoms Embeddings with Cough Embeddings followed by a FC layer.
\begin{equation}
    \hat{y} =   \underbrace{\textstyle{\bigl[S_{e}, C_{e}\bigr]}}%
    _{\text{Concatenate}} \cdot FC
\end{equation}
\figureref{fig:maindiagram} shows the overall structure of the proposed architecture.
After this, Total loss was calculated as follows

\begin{equation}
Loss_{total}=\underbrace{\textstyle(1-\alpha)\; Loss_{ce}}_{\substack{\text{ \tiny Cough} \\ \text{ \tiny Embeddings Loss}}}+ \underbrace{\textstyle \alpha\; Loss_{se}}_{\substack{\text{ \tiny Symptoms} \\ \text{\tiny Embeddings Loss}}}
\end{equation}

Where $\alpha$ is a small constant value to balance the contribution of the different losses.
\section{Experiments}
\subsection{Evaluation}
In this section, a comprehensive evaluation is carried out to investigate the results of four clinical classification tasks. Based on the dataset collected, the model was trained on the following combination of features.
\begin{itemize}
\item  \textbf{Task 1, Using cough data only} In this experiment setup, only cough features were utilized from the collected dataset to train the Model and distinguish between COVID-19 positive and negative cases. Cough features were extracted using the signal processing pipeline, as described in section 1.
\begin{table*}
    \small
\centering
\begin{tabular}{c | c|ccccc}
    \toprule
\cmidrule(lr){5-7}&  &\textbf{F1-score}  &  \textbf{Precision}  & \textbf{Sensitivity}  & \textbf{Specificity}  & \textbf{Accuracy}          \\
    \midrule
    \multirow{4}{*} {\footnotesize \textbf{Cough data} }
        & \footnotesize Covid-19 Positive    & \footnotesize 90.6 $\pm$ 0.2\% &  \footnotesize 89.1 $\pm$ 0.4\% & \footnotesize 86.2 $\pm$ 0.3\% & \footnotesize 92.4 $\pm$ 0.2\%& \footnotesize 89.3 $\pm$ 0.1\%\\
        & \footnotesize Covid-19 Negative    & \footnotesize 90.6 $\pm$ 0.1\% & \footnotesize 91.7 $\pm$ 0.1\% & \footnotesize 94.3 $\pm$ 0.3\%& \footnotesize 89.3 $\pm$ 0.1\% &\footnotesize 92.4 $\pm$ 0.1\%            \\
        \cmidrule{2-7}
        & \footnotesize Overall & \footnotesize 90.6 $\pm$ 0.3\% & \footnotesize 90.4 $\pm$ 0.5\% & \footnotesize 90.1 $\pm$ 0.6\% & \footnotesize 90.3 $\pm$ 0.3\% & \footnotesize 90.8 $\pm$ 0.2\%
        \\
        \cmidrule{2-7}
\multirow{4}{*}{\footnotesize \textbf{Symptoms data} }
       & \footnotesize Covid-19 Positive    & \footnotesize 91.5 $\pm$ 0.2\% & \footnotesize 86.9 $\pm$ 0.5\% & \footnotesize 87.8 $\pm$ 0.2\% & \footnotesize 86.0 $\pm$ 0.3\% & \footnotesize 94.1 $\pm$ 0.6\%           \\
        & \footnotesize Covid-19 Negative    & \footnotesize 91.5 $\pm$ 0.1\% & \footnotesize 93.7 $\pm$ 0.3\% & \footnotesize 93.3 $\pm$ 0.2\% & \footnotesize 94.1 $\pm$ 0.1\% & \footnotesize 86.0 $\pm$ 0.2\%\\
        \cmidrule{2-7}
        & \footnotesize Overall   & \footnotesize 91.5 $\pm$ 0.3\%& \footnotesize 90.3 $\pm$ 0.8\%& \footnotesize 90.5 $\pm$ 0.4\%& \footnotesize 90.8 $\pm$ 0.3\%&\footnotesize  91.1 $\pm$ 0.8\%\\
    \cmidrule{2-7}
    \multirow{4}{*}{\footnotesize \textbf{Both} }
       & \footnotesize Covid-19 Positive    & \footnotesize 96.8  $\pm$ 0.4\% &\footnotesize 95.1 $\pm$ 0.1\%&\footnotesize 94.6 $\pm$ 0.3\%&\footnotesize 95.6 $\pm$ 0.1\%&\footnotesize 97.3 $\pm$ 0.2\%           \\
        & \footnotesize Covid-19 Negative    & \footnotesize 96.8 $\pm$ 0.1\% &\footnotesize 97.6 $\pm$ 0.3\% & \footnotesize 97.8 $\pm$ 0.4\% &\footnotesize 97.3 $\pm$ 0.2\%& \footnotesize 95.6 $\pm$ 0.3\% \\
        \cmidrule{2-7}
        & \footnotesize Overall & \footnotesize 96.8 $\pm$ 0.5\% &\footnotesize 96.3 $\pm$ 0.4\%&\footnotesize 96.2 $\pm$ 0.7\%&\footnotesize 96.5 $\pm$ 0.3\%&\footnotesize 96.5 $\pm$ 0.5\%
        \\
    \bottomrule
\end{tabular}
\caption{ Model performance metrics across different models on Covid-19 data }
\label{tab:allcls}
    \end{table*}
    
\begin{figure*}[ht]
\centering
  
  \begin{minipage}[b]{0.43\linewidth}
    \centering
    \includegraphics[width=0.80\linewidth]{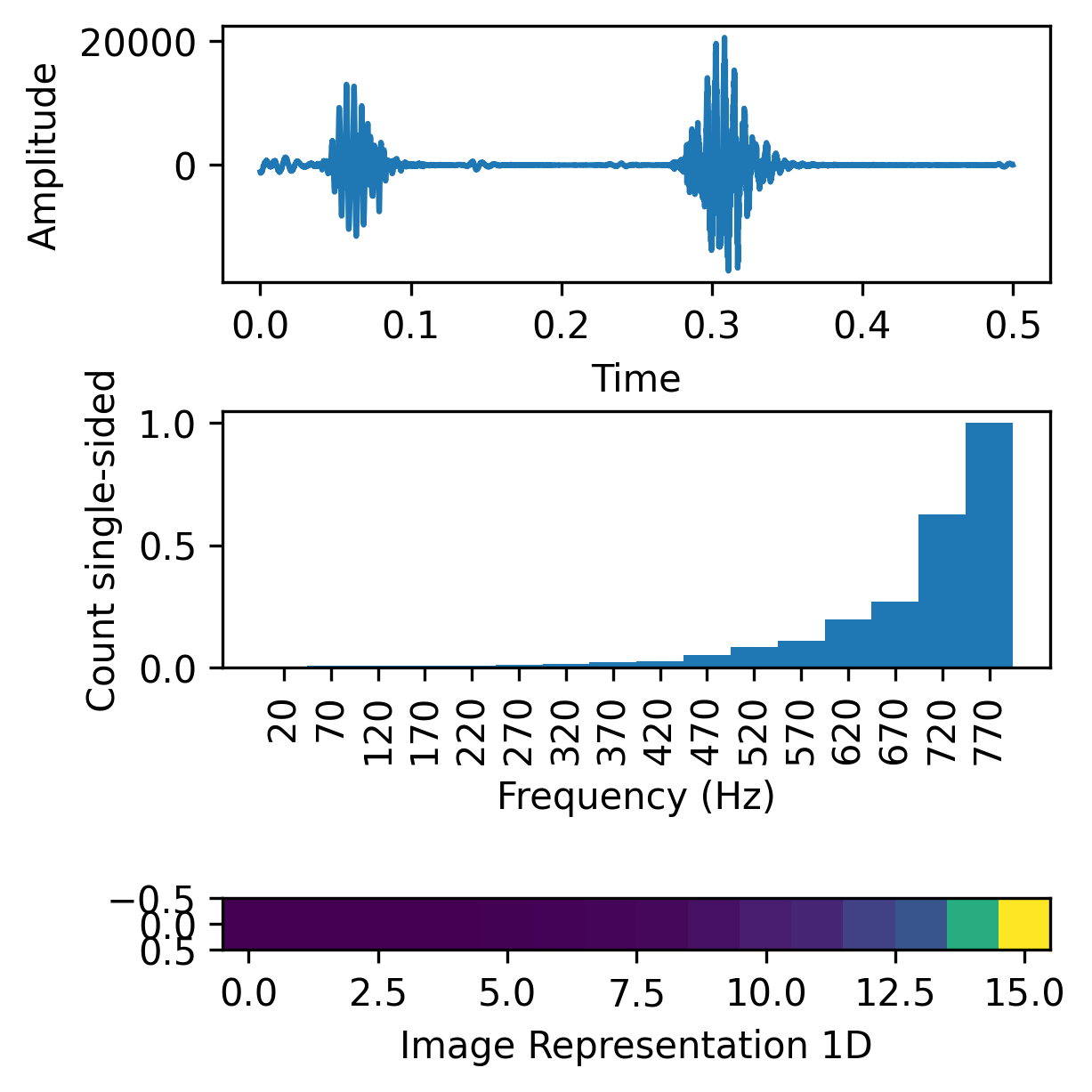} 
    \caption{\footnotesize Healthy Cough} 
    \label{fig:stasdet0}
    \vspace{3ex}
  \end{minipage}
  \begin{minipage}[b]{0.43\linewidth}
    \centering
    \includegraphics[width=.80\linewidth]{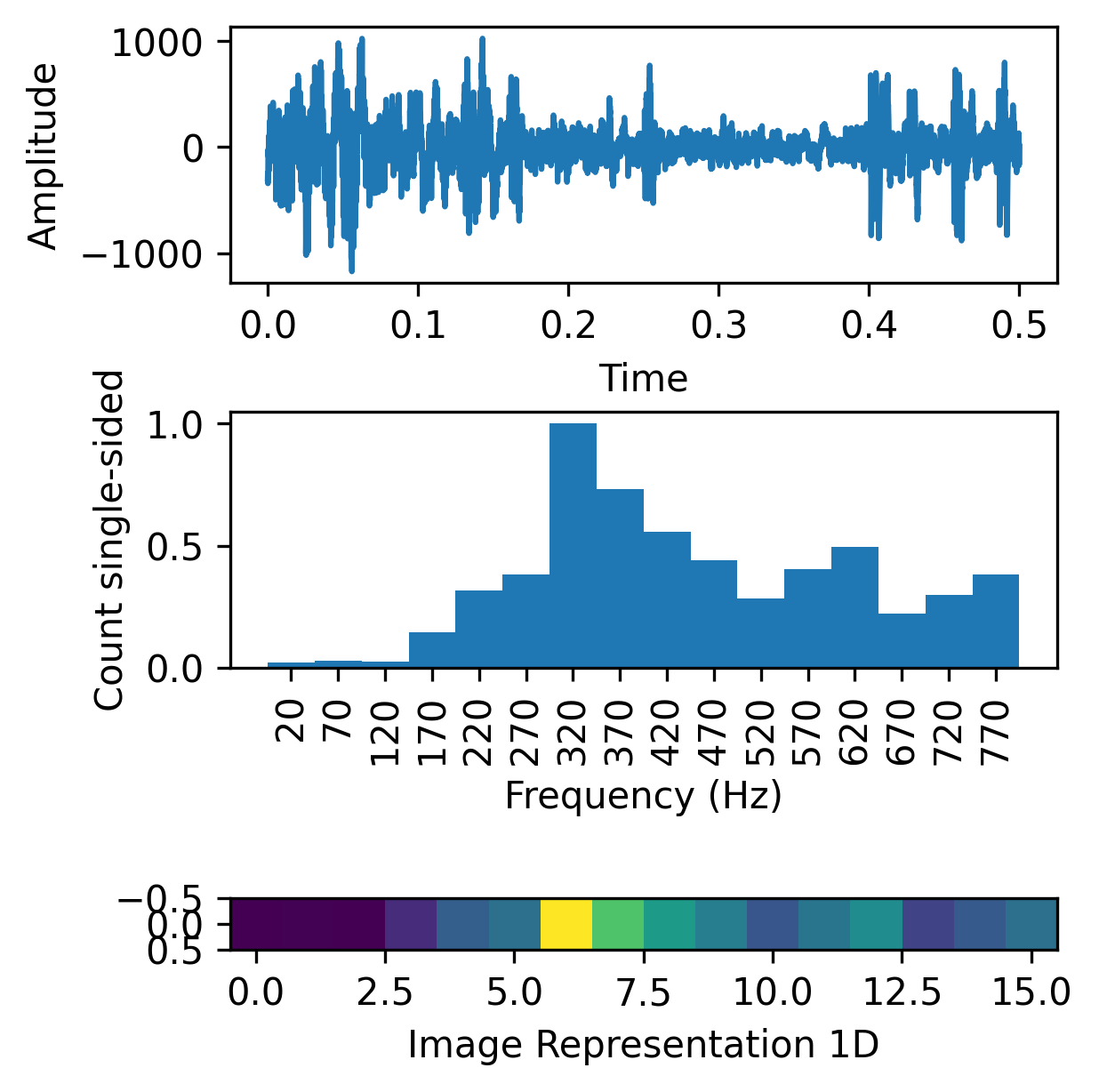} 
    \caption{\footnotesize Asthma Cough} 
    \label{fig:stasdet1}
    \vspace{3ex}
  \end{minipage} 
  \begin{minipage}[b]{0.43\linewidth}
    \centering
    \includegraphics[width=.80\linewidth]{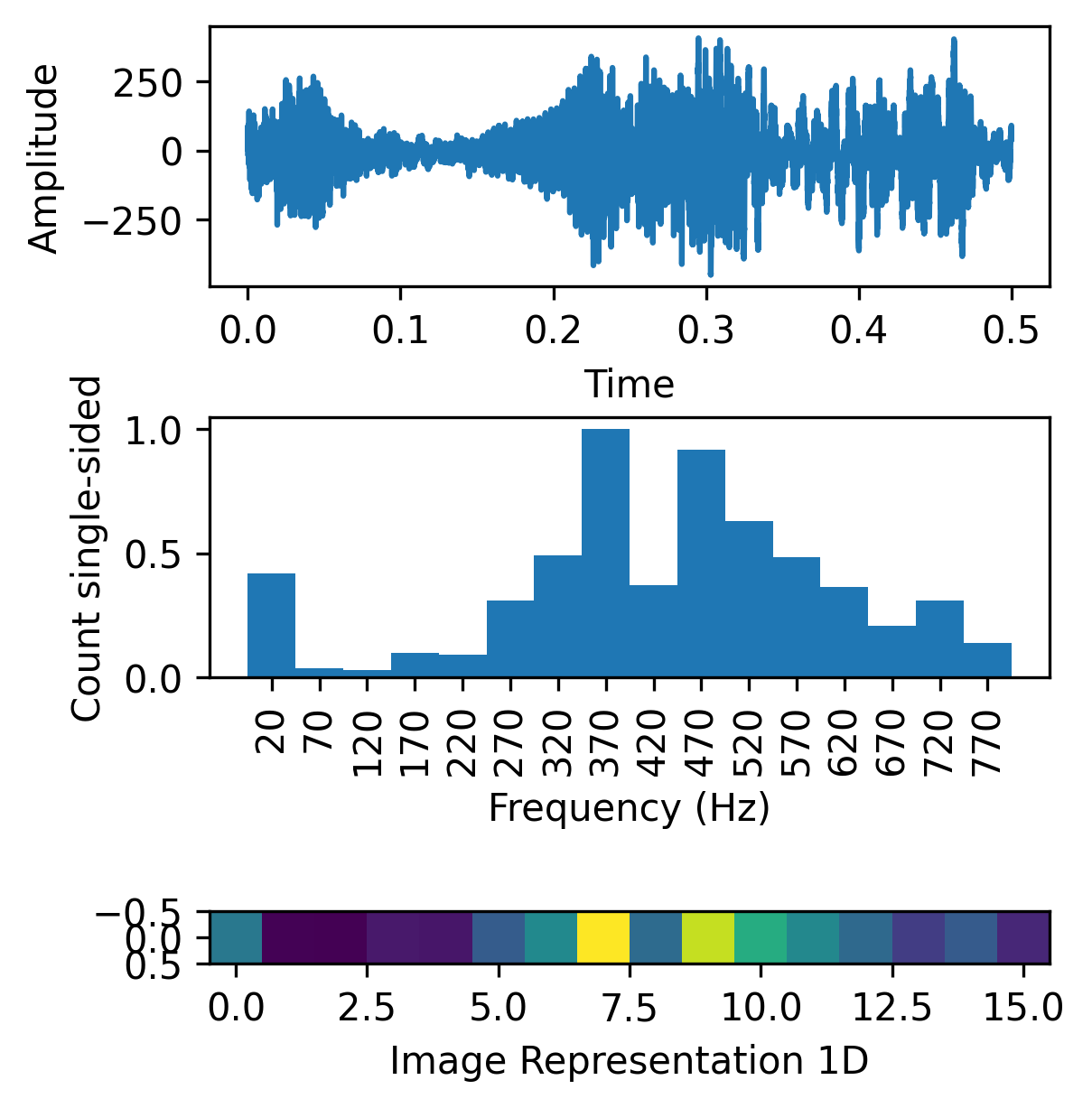} 
    \caption{\footnotesize Bronchitis Cough} 
    \label{fig:stasdet2}
    \vspace{3ex}
  \end{minipage}
  \begin{minipage}[b]{0.43\linewidth}
    \centering
    \includegraphics[width=.80\linewidth]{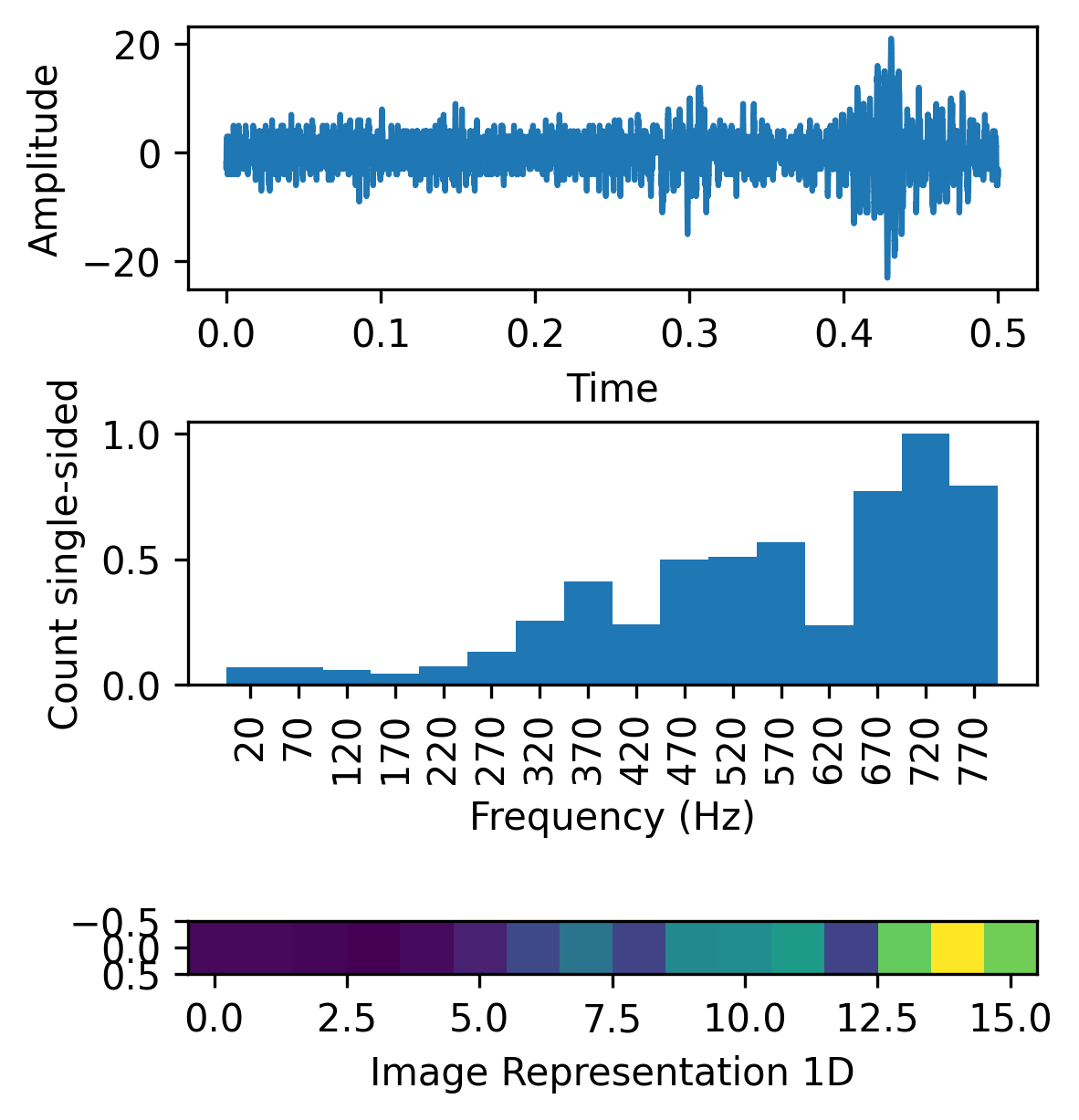} 
    \caption{\footnotesize COVID-19 Cough}
    \label{fig:stasdet3}
    \vspace{3ex}
  \end{minipage}
  \caption{ \footnotesize Four types of cough with their original sound, FFT output, and 1D image representation in Figure 2: Healthy cough, Figure 3: Asthma Cough, Figure 4: Bronchitis cough, and Figure 5: Covid-19 positive cough }
  \label{fig:stasdet}
\end{figure*}

\begin{figure*}[htb]
  \centering
  \includegraphics[width=12cm]{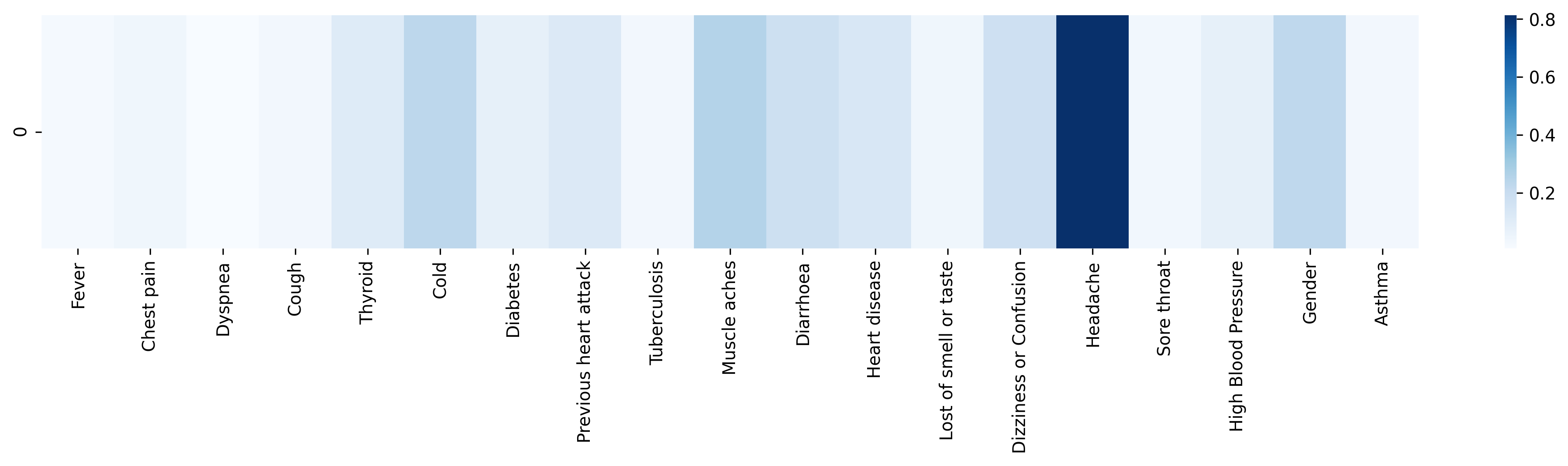}
  \caption { \footnotesize Attention distribution over the Symptoms of a Healthy(COVID-19 Negative) person. The color depth expresses the seriousness of a symptom.}
  \label{fig:att0}
  \vspace{0.5\baselineskip}
  \includegraphics[width=12cm]{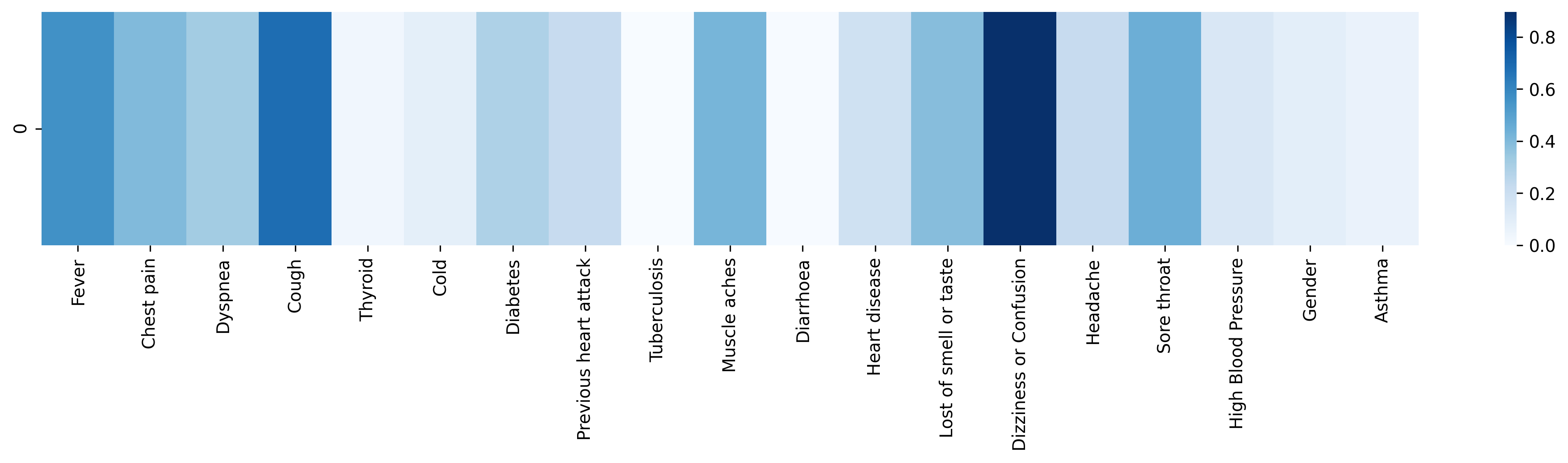}
  \caption { \footnotesize Attention distribution over the Symptoms of a COVID-19 infected person. Fever, Cough, Dizziness, or Confusion, Chest pain is with high color depth, showing that the model has learned the symptoms embedding based on demographic \& cough features.}
  \label{fig:att1}
\end{figure*}
\begin{figure}[h]
\includegraphics[width=7.3cm]{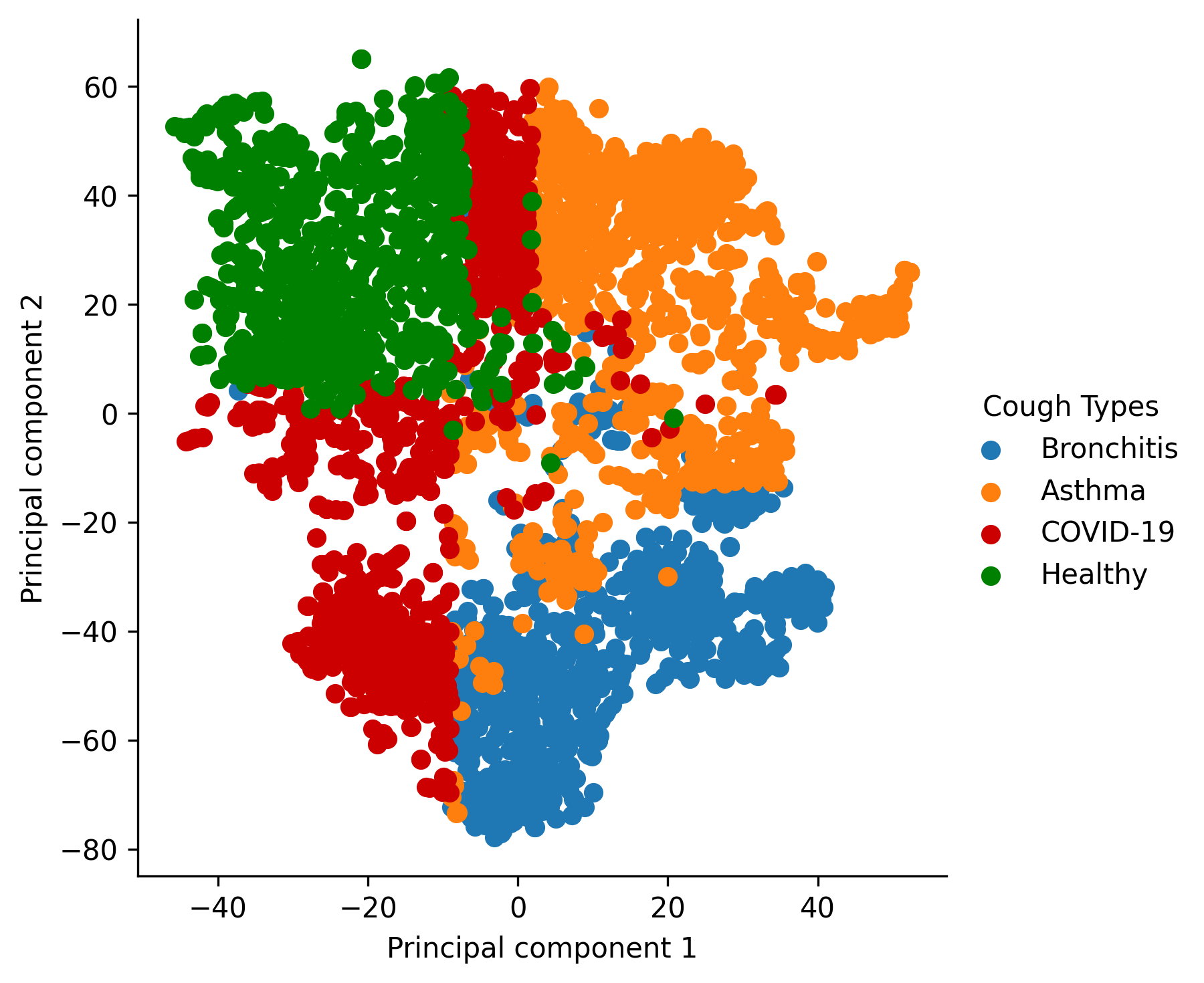}
\caption{ \footnotesize t-SNE visualization of four types of Cough features }
\label{fig:att2}
\end{figure}

\item \textbf{Task 2, Using Demographic \& Symptoms Data Only} In this setup, experiments were conducted on Demographic \& Symptoms data. The symptoms ( Fever, Headache, aches, sore throat, etc.) were used to train the model and classify between cover-19 positive and negative cases.
\item \textbf{Task 3, Using Both} when using both types of data, including cough features from section 3 with Demographic \& Symptoms data, Model learn the hidden patterns and relationship between both types of features and classify between COVID-19 positive and negative cases. 

\item \textbf{Task 4, Using Both with different Cough Types}, to demonstrate the effectiveness of the Model, the Model is trained on four different types of cough, including COVID-19, Bronchitis, Asthma, and Healthy using both types of data.
\end{itemize}
The five standard evaluation metrics (Accuracy, specificity, sensitivity/recall,  precision, and F1-score) are adopted to evaluate the Model on the test dataset. Several iterations were performed, and results were reported. \tableref{tab:fourdiseases} and \tableref{tab:allcls} represents the classification of results along with the following observations:

\begin{itemize}
 \item The Model achieves Accuracy, Specificity, and Sensitivity of 90.8 $\pm$ 0.2\%, 90.3 $\pm$ 0.3\%, 90.1 $\pm$ 0.6\% respectively. Task 1, using only the cough features. It demonstrates that Cough provides requisite data about the respiratory system and the pathogenies involved. Signal processing characteristics enable the Model to capture the hidden cough sound signatures and diagnose COVID-19 with sufficient sensitivity and specificity.
\item When using symptoms and demographic data, the Model's performance represents a slight increase in Accuracy, Specificity, and Sensitivity 91.1 $\pm$ 0.8\%, 90.8 $\pm$ 0.3\%, and 90.5 $\pm$ 0.4\%. Compared to Task 1. This increase is attributed to the rich categorical information present in Demographic and Symptoms data about suspected infection cases and Transformer based network; TabNet exploits attention mechanism to learn the characteristics, diagnosis automatically, and symptoms based on information in their dataset.
\item In Task 3, the Model makes better use of available data by combining both types of representation, which complement each other and significantly improve the classification's performance with the Accuracy, Specificity, and Sensitivity of 96.5 $\pm$ 0.5\%, 96.5 $\pm$ 0.3\% and 96.2 $\pm$ 0.7\%. It is observed that one data type, either cough features or symptoms' features, is insufficient to capture the features predicting COVID-19 disease effectively.
\item To further demonstrate the proposed Model's flexibility, the Model was trained with a multi-class classification setting to distinguish between four types of cough classes, including Bronchitis, Asthma, COVID-19 Positive, and COVID-19 Negative. The experimental results of Multi-class classification are presented in \tableref{tab:fourdiseases}. Results show that the Model can certainly classify between four types of cough classes with Accuracy, Specificity, and Sensitivity of 95.04 $\pm$ 96.83 $\pm$ 0.06\%,90.41 $\pm$ 0.14\% respectively. 

Both types of data enable the Model to learn deeper relationships between temporal acoustic characteristics of cough sounds and Symptoms' features and hence perform better.
\end{itemize}

\subsection{Interpretability}
It is demonstrated that the proposed framework benefits from the high accuracy and generality of deep neural networks and TabNet's interpretability, which is crucial for AI-empowered healthcare. \figureref{fig:att0} and \figureref{fig:att1}  visualizes the symptoms of a healthy and COVID-19 infected individual. It shows that the model comprehends the hidden pattern in symptoms data and its relationship with cough sounds. To intuitively show the representation's quality, the cough features using t-sne and symptoms correlation matrix are visualized in \figureref{fig:att2} and \figureref{fig:corr}

\subsection{In Depth Clinical Analysis}
An in-depth analysis is conducted for different cough sounds diagnosed with different diseases based on the collected data.
Different types of cough samples are visualized in \figureref{fig:stasdet}. Based on the analyzed data, the findings are as follows.

The coughing sound consists of three phases- Phase 1- Initial burst, Phase 2- Noisy airflow, and Phase 3- Glottal closure. It is observed that in the cough sample of healthy individuals, phase 3 finished with vocal folds activity. \figureref{fig:stasdet0} shows that after Phase 1, i.e., initial burst, the energy levels are high at higher frequencies.

Asthma \& Bronchitis comes under the wet cough (carries mucus and sputum caused by bacteria or viruses, secretion in the lower airways) category. In particular, vocal fold activity looks random, and the energy is expended over a broader frequency band. Asthma \& Bronchitis \figureref{fig:stasdet1} and \figureref{fig:stasdet2} shows these characteristics and the energy level.

It is observed that COVID-19 cough is continuous; energy distribution is spread across frequencies preceded by a short catch. By analyzing the mean energy distribution of many COVID-19 cough sounds, Energy distribution was high in Phase 2 and Phase 3. The abnormal oscillatory motion in the vocal folds may be produced by altered aerodynamics over the glottis due to respiratory irritation. \figureref{fig:stasdet3} shows the result
\section{Conclusion}
Mass COVID-19 monitoring has proved essential for governments to successfully track the disease's spread, isolate infected individuals, and effectively "flatten the curve" of the infection over time. In the wake of the COVID-19 pandemic, many countries cannot conduct rapid enough tests; hence an alternative could prove very useful. This study brings forth a Low cost, accurate and interpretable AI-based diagnostic tool for COVID-19 screening by incorporating the demographic, symptoms, and cough features and achieving mean accuracy, precision, and precision in the mentioned tasks. This significant achievement supports large-scale COVID-19 disease screening and areas where healthcare facilities are not easily accessible. Data collection is being performed daily. Experiments will be carried out in the future by incorporating different voice data features such as breathing sound, counting sound (natural voice samples), and sustained vowel phonation.
The results prove to be transparent, interpretable, and multi-model learning in cough classification research.

\appendix

\section{Cough Features}\label{apd:cfeatures}

\subsection{Mel Frequency Cepstral coefficients (MFCCs)}
Mel Frequency Cepstral coefficients represent the short-term power spectrum of a signal on the Mel-scale of the frequency. The hearing mechanism of human beings inspires MFCC. The coefficients of MEL-frequency that represent this transformation are called  MFCCs. 
First, we apply the Discrete Fourier Transform(DFT) on each cough sub-segment 

\begin{multline}
Y(k)=\smash[b]{\sum_{t=0}^{N-1}} y_{i}[t] w(t) \exp(-2 \pi i kt/N), \\
 k=0,1, \ldots, N-1 \label{eq:dft}
\end{multline}
Where $N$ denotes the number of samples in  frame, $y_{i}[t]$ is the discrete time  domain cough signal, obtained in Section~\ref{sec:sp}, $w(t)$ is the window function in time domain and $Y(k)$ is the $k^{t h}$ harmonic corresponding to the frequency $f(k)=k F_{s} / N$ \; where $F_{s}$ is the sampling frequency.
MFCCs use Mel filter bank or triangular bandpass filter on each cough signal's DFT output, equally spaced on the Mel-scale. 
At last, we apply the Discrete Cosine Transform(DCT) on the output of the log filter bank in order to get the MFCCs :

\begin{equation}
c(i)=\sqrt{\frac{2}{M}} \sum_{m=1}^{M} \log (E(m)) \cos \left(\frac{\pi i}{M}(m-0.5)\right)
\end{equation}

where $i=1,2, \ldots, l, l$ denotes the cepstrum order, $E(m)$ and $M$ are the filter bank energies and total number of mel-filters respectively. 

\subsection{Log energy}
To calculate the log energy of each sub-segment, the following formula was used:

\begin{equation}
\mathrm{L}_{t}=10 \log 10\left(\varepsilon+\frac{1}{N} \sum_{t=1}^{N} y_{i}(t)^{2}\right)
\end{equation}

where $\varepsilon$ is a minimal positive value.
\subsection{Zero crossing rate(ZCR)}
ZCR is used to calculate the number of times a signal crosses the zero axis. To detect the cough signal's periodic nature, we compute the number of zero crossings for each sub-segment.
\begin{equation}
Z_{t}=\frac{1}{N-1} \sum_{t=1}^{N-1} \Pi\left[y_{i}(t) y_{i}(t-1)\right]
\end{equation}
Where $\Pi[A]$ is a indicator function and is defined as
\begin{align*}
  \Pi[A] = \left \{
  \begin{aligned}
    &1, && \text{if}\ A < 0 \\
    &0, && \text{otherwise}
  \end{aligned} \right.
\end{align*}

\subsection{Skewness}
Skewness is the third order moment of a signal, Which measures the symmetry in a probability distribution.

\begin{equation}
\mathrm{S}_{t} =\frac{E\left(y_{i}[t]-\mu\right)^{3}}{\sigma^{3}}
\end{equation}

Where $\mu$ and $\sigma$ is mean and stand deviation of the sub-segment $y_{i}[t]$ respectively.
\subsection{Entropy}
We compute the Entropy for each sub-segment of the cough signal to capture the difference between signal energy distributions.
\begin{equation}
\mathrm{E}_{t}=-\sum_{t=1}^{N-1}y_{i}(t)^{2} \ln (y_{i}(t)^{2}), \quad 1 \leq t \leq N-1
\end{equation}
\begin{figure*}
  \includegraphics[width=16cm]{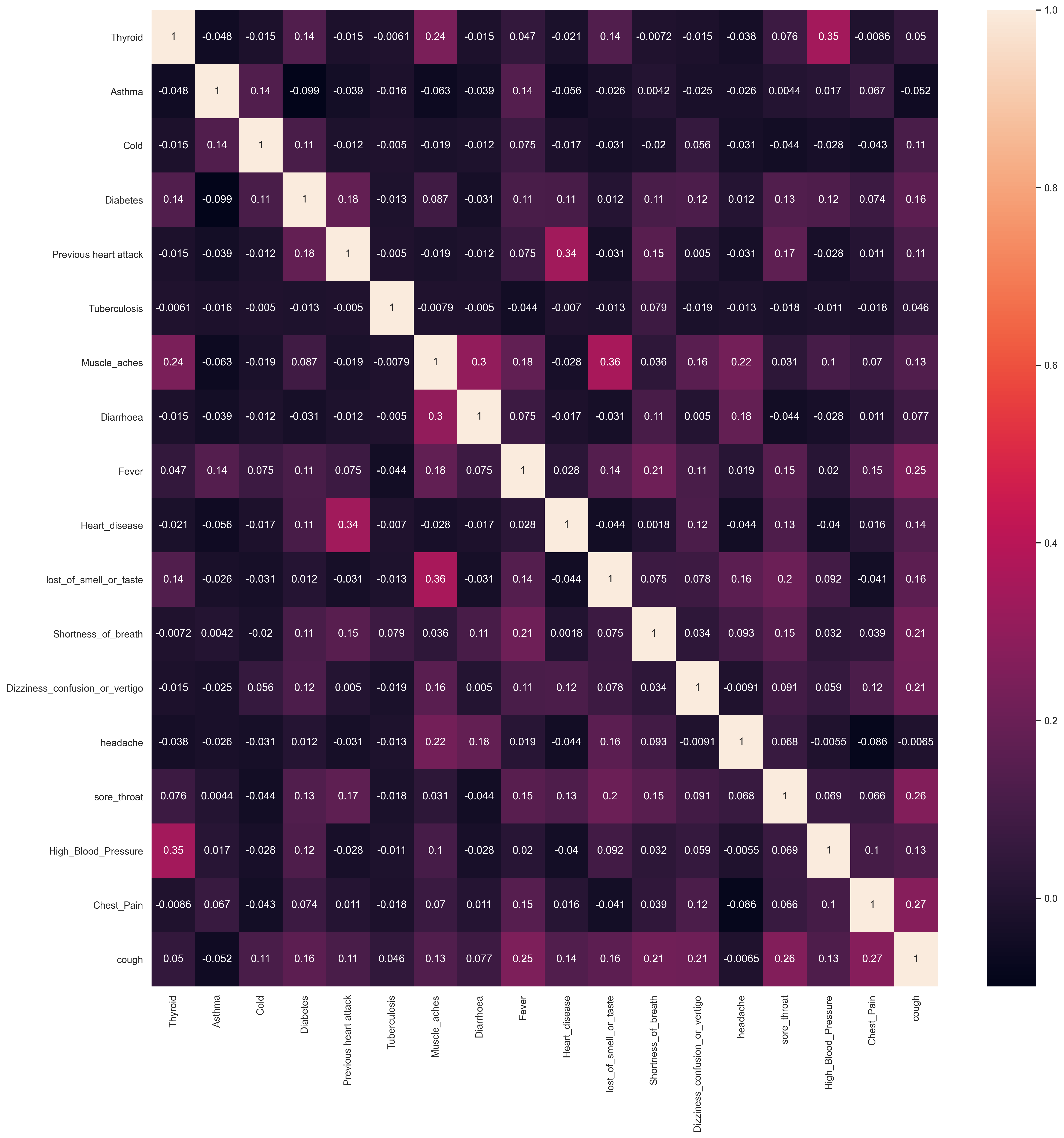}
  \caption{\footnotesize Correlation matrix for Symptoms and medical conditions data}
  \label{fig:corr}
\end{figure*}

\subsection{Formant frequencies}
In the analysis of speech signals,  Formant frequencies are used to capture a human vocal tract resonance's characteristics. We compute the Formant frequencies by peak picking the Linear Predictive Coding(LPC) spectrum. We used the Levinson-Durbin recursive procedure to select the parameters for the 14th order LPC model. The first four Formant frequencies(F1-F4) are enough to discriminate various acoustic features of airways. 

\subsection{Kurtosis}
Kurtosis can be defined as the fourth-order moment of a signal, which measures the peakiness or heaviness associated with the cough sub-segment probability distribution.

\begin{equation}
\mathrm{k}_{t} =\frac{E\left(y_{i}[t]-\mu\right)^{4}}{\sigma^{4}}
\end{equation}

Where $\mu$ and $\sigma$ is mean and stand deviation of the sub-segment $y_{i}[t]$ respectively.

\subsection{Fundamental frequency(F0)}
To estimate the fundamental frequency (F0)  of the cough sub-segment, we used the center-clipped auto-correlation method by removing the formant structure from the auto-correlation of the cough signal.

\section{TabNet Components} \label{apd:tabf}

\subsubsection{Gated Linear Unit (GLU)} \label{apd:gll}
GLU \citep{dauphin2017language} layer consist of Fully connected layer (FC), Ghost batch normalization (GBN) \citep{hoffer2018train} and GLU. FC layer map the input features to 2($n_{d}$ + $n_{a}$) where $n_{d}$ and $n_{a}$ are size of the decision layer and size of the attention bottleneck respectively. GBN splits each batch in chunks of virtual batch size(BS), standard Batch normalization(BN) applies separately to each of these,  and concatenates the results back into the original batch.

\subsubsection{Feature Transformer (FT)}\label{apd:ft}
The feature transformer is one of the main components of TabNet; it consists of 4 GLU layers, two are shared across the entire network, and two are step-dependent across each decision step allowing for more modeling flexibility.  GLU layers are concatenated with each other after being multiplied by a constant scaling factor($\sqrt{0.5}$). Feature Transformer process the filtered features by looking at all the symptoms features assessed and deciding which ones indicate which class.

\subsubsection{Attentive Transformer (AT)}\label{apd:at}
Attention Transformer is another main component of TabNet architecture. It utilizes sparse intense wise features selection based on learned symptoms dataset and directs the model’s attention by forcing the sparsity into the feature set, focusing on specific symptoms features only.
It is a powerful way of prioritizing which features to look at for each decision step. The FC handles the learning part of this block. TabNet uses the Sparsemax, \cite{martins2016softmax} an alternative of softmax function for soft feature selection. 
Sparsemax activation function is differentiable and has forward and backward propagation. Due to projection and thresholding, sparsemax process sparse probabilities lead to a selective and more compact attention focus on symptoms features.
\section{Data Acquisition}\label{apd:dataacq}

All the COVID-19 data utilized in this study were obtained from 200 subjects in a Dr. Ram Manohar Lohia Hospital, New Delhi, India. Out of 100 were confirmed positive from COVID-19 reverse transcription-polymerase chain reaction (RT-PCR) results. The Clinical Trials Registry-India (CTRI) had approved the study protocols and the patient recruitment procedure. After data preprocessing and Out of 200 samples, 50 samples were discarded due to low data quality.
Aside from COVID-19 and healthy data, We also collected Bronchitis and Asthma cough from different online and offline sources. The data collection person followed all the clinical safety measures and inclusion-exclusion criteria and the cough sounds, breathing sounds, counting 1 to 10 ( natural voice samples ), sustained phonation of 'a,' 'e,' 'o' vowel, demographic, symptoms data such as fever, headache, sore Thorat, or any other medical conditions were also collected at the same time. The average interaction time with the subject was 10--12 mins.

\bibliography{jmlr-sample}

\end{document}